\documentclass[useAMS,referee,usenatbib]{biom}

\def\bSig\mathbf{\Sigma}

\usepackage[figuresright]{rotating}
\usepackage{siunitx}
\usepackage{verbatim}
\usepackage{float}
\usepackage{mathrsfs}
\usepackage{longtable}
\usepackage{tabularx}
\usepackage{rotating}
\usepackage{amsfonts}
\usepackage{amsmath}
\usepackage{hhline}
\usepackage{xcolor}
\usepackage{url}
\newcolumntype{C}[1]{>{\centering\arraybackslash}p{#1}}
\title[Analyses and Concerns in Precision Medicine: A Statistical Perspective]{Analyses and Concerns in Precision Medicine: A Statistical Perspective}

\author{Xiaofei Chen$^{1}$\email{xiaofeic@smu.edu}\\
$^{1}$ Department of Statistical Science, Southern Methodist University, Dallas, TX, USA\\
}

\begin{document}
\label{firstpage}

\begin{abstract}
This article explores the critical role of statistical analysis in precision medicine. It discusses how personalized healthcare is enhanced by statistical methods that interpret complex, multidimensional datasets, focusing on predictive modeling, machine learning algorithms, and data visualization techniques. The paper addresses challenges in data integration and interpretation, particularly with diverse data sources like electronic health records (EHRs) and genomic data. It also delves into ethical considerations such as patient privacy and data security. In addition, the paper highlights the evolution of statistical analysis in medicine, core statistical methodologies in precision medicine, and future directions in the field, emphasizing the integration of artificial intelligence (AI) and machine learning (ML).
\end{abstract}


\maketitle
\section{Introduction}
Precision medicine represents a trans-formative approach to healthcare, shifting the paradigm from a one-size-fits-all strategy to one that is tailored to individual patients based on their genetic makeup, environment, and lifestyle \citep{adam2007personalised}. This personalized approach not only enhances the efficacy of treatments but also minimizes the risk of adverse effects \citep{agyeman2015perspective, kumari2023advancements}. However, the success of precision medicine heavily relies on the interpretation of complex, multidimensional data sets, where statistical analysis plays a pivotal role \citep{alyass2015big}.
The integration of statistical methodologies in precision medicine is not just a mere addition but a fundamental necessity. Advanced statistical techniques enable the extraction of meaningful insights from large-scale genomics, proteomic, and metabolomic data, which are the cornerstone of precision medicine \citep{wafi2018translational, pinu2019translational}. These methodologies include, but are not limited to, predictive modeling, machine learning algorithms, and complex data visualization techniques, all of which contribute to more accurate diagnosis, prognosis, and treatment planning \citep{bellazzi2008predictive, davatzikos2018cancer, richter2018review}.

The heterogeneity of data sources in precision medicine, ranging from electronic health records (EHRs) to high-throughput sequencing data, presents unique challenges in data integration and interpretation \citep{martinez2022data}. Statistical analysis serves as a bridge, merging these diverse data types into coherent, interpretable information that can guide clinical decision-making.

However, the field is not without its challenges. Issues such as overfitting, handling of high-dimensional data, and maintaining the balance between model complexity and interpretability are ongoing areas of research \citep{bolon2015feature, xu2019review, bommert2020integration, pes2020learning, hou2022dimensionality}. In addition, ethical considerations, such as patient privacy and data security, are intertwined with statistical analysis, necessitating a careful, well-thought-out approach \citep{miles2005handbook, vaught2011biological, elliot2020anonymisation}.

In this review, we will explore the various statistical methods employed in precision medicine, their applications, and the challenges they present. We aim to provide a comprehensive overview of how statistical analysis is shaping the future of personalized healthcare and the implications for patients, healthcare providers, and researchers.

\section{Evolution of Statistical Analysis in Precision Medicine}
The roots of statistical analysis in medicine can be traced back to the early 20th century, with the pioneering work of statisticians like Ronald Fisher and Karl Pearson \citep{rao1992ra, stigler2002statistics, ougucs2017together}. Their work laid the foundation for the application of statistical methods in biological and medical research. However, it wasn't until the late 20th century, with the advent of more sophisticated computational tools and the Human Genome Project (HGP), that statistical analysis began to play a crucial role in medicine \citep{collins2001implications,hood2013human}.

The completion of the Human Genome Project (HGP) in 2003 marked a significant milestone \citep{zwart2015human}. It paved the way for the development of genomics medicine, necessitating advanced statistical tools to analyze the vast amounts of genetic data. Subsequently, the introduction of technologies like high-throughput sequencing and bioinformatics tools further revolutionized the field \citep{delseny2010high, qiang2014high}. These advancements enabled the identification of genetic markers for diseases and the development of targeted therapies \citep{li2020applications}.

Traditional statistical methods in medicine focused primarily on hypothesis testing and linear models, which were suitable for smaller, simpler datasets \citep{zhou2009statistical}. However, the complexity and volume of data in precision medicine required more sophisticated approaches. This led to the adoption of machine learning algorithms, Bayesian methods, and network analysis, which are better suited for handling high-dimensional data and identifying complex patterns \citep{ivanovic2015modern, sarker2021machine}. These techniques have allowed for more accurate predictions of disease risk, drug response, and treatment outcomes, marking a significant evolution in the statistical approach to medicine.

\section{Core Statistical Methodologies in Precision Medicine}
\subsection{Traditional Hypotheses Testing Methods}
Traditional hypothesis testing methods are foundational to statistical analysis and scientific research. They help determine if there is enough evidence to support a specific claim or hypothesis. The most common methods include: T-test, Chi-Sqaure test, ANOVA (Analysis of Variance), etc. 

The t-test is a statistical test widely used to compare the means of two groups, and it's especially useful when dealing with small sample sizes \citep{kim2015t}. It helps determine if there are significant differences between the means of two groups, which may be related in certain features \citep{de2019using}. The t-test is a versatile statistical tool with wide-ranging applications in various fields \citep{feng2017application,chen2023sex}. 

The Chi-Square Test, often represented as $\chi^2$ test, is a statistical method used to determine if there is a significant association between two categorical variables. It's widely used in various fields like research \citep{wang2017strong}, business \citep{mirajkar2013application}, healthcare, and social sciences \citep{onchiri2013conceptual}. The Chi-Square Test is particularly useful for handling data that is in nominal or categorical form.
It allows researchers to understand and interpret the relationships between categorical variables, such as the link between genetic traits or the association between environmental factors and species distribution.

Analysis of Variance (ANOVA) is a statistical method used to compare the means of three or more independent groups to see if at least one group mean is significantly different from the others. It's an extension of the t-test, which is typically used for comparing means of two groups. ANOVA is particularly useful when dealing with multiple groups and variables, making it a staple in various fields such as psychology, medicine, business, and biological sciences \citep{christensen1996analysis, gelman2005analysis, sawyer2009analysis}.

\subsection{Predictive Modeling}
Predictive modeling is a statistical technique used to forecast outcomes or behaviors by analyzing current and historical data. It encompasses a variety of statistical models and algorithms that are used to identify patterns and relationships in data, which can then be used to make predictions about future or unknown events. Some key predictive modeling techniques: linear regression, logistic regression, survival analysis, semi-parametric predictive models, etc.

Linear regression is one of the most fundamental and widely used statistical techniques for predictive modeling and data analysis \citep{weisberg2005applied, yuan2018subgroup}. It's a linear approach to modeling the relationship between a dependent variable and one or more independent variables. The method of least squares is used to find the line that minimizes the sum of squared residuals (differences between observed and predicted values). The coefficients represent the change in the dependent variable for a one-unit change in the independent variable, holding other variables constant \citep{montgomery2021introduction}.

Linear regression is a versatile tool used across various domains. In economics and finance, it helps in market analysis, such as predicting stock prices and assessing credit risk, and in developing pricing models for commodities and real estate. Businesses use linear regression for sales forecasting and to measure the effectiveness of marketing campaigns. In healthcare, it's applied in medical research to estimate treatment effects and in predicting healthcare costs. Environmental scientists use it to model the impacts of environmental factors on climate indicators and to study ecological relationships. Social scientists apply linear regression to explore factors affecting outcomes in education and sociology, such as student performance and employment rates. In the real estate sector, it's used for property price estimation and understanding market trends. Each of these applications leverages the ability of linear regression to uncover relationships between variables and make predictions based on historical data.

Logistic regression, a type of generalized linear regression, is particularly tailored for binary classification problems \citep{lavalley2008logistic}. Unlike linear regression that predicts a continuous outcome, logistic regression is used when the outcome is categorical, typically in a binary format like Yes/No, True/False, or 1/0. It estimates the probability that a given input point belongs to a certain class. The core mechanism involves the logistic function, which bounds the output of the model between 0 and 1, making it interpretable as a probability.

In medical research, logistic regression has a wide array of applications \citep{zhang2016visual}. It's frequently used for disease prediction, where the outcome could be the presence or absence of a disease based on a set of predictors like age, weight, genetics, and lifestyle factors. For instance, it might be used to calculate the likelihood of a patient developing diabetes \citep{wilson2007prediction} based on their insulin levels, diet, and body mass index. Logistic regression is also valuable in analyzing risk factors for diseases, enabling researchers to identify which factors are significant predictors of certain medical conditions or predict the substances use in youth \citep{loffredo2017prevalence}. This method is crucial in decision-making for preventive medicine, where understanding the probability of disease occurrence can guide medical professionals in recommending lifestyle changes or interventions. 

Survival analysis is a branch of statistics that deals with the analysis of time-to-event data, commonly referred to as survival data. The term 'survival' stems from its origins in clinical research, where it was used to determine the time until an event of interest, like death or relapse, occurs. However, its applications extend far beyond medical research. Survival analysis is distinct in its handling of censored data--situations where the exact time of the event is not known for all subjects, either because the event has not yet happened or because the observation period ended. This feature makes it especially useful for analyzing lifespan or failure time data in a wide range of contexts. The techniques of survival analysis, such as the Kaplan-Meier estimator and the Cox proportional hazards model, allow for the estimation of survival probabilities over time and the examination of the effects of predictor variables on the survival experience. For example, in medical research, survival analysis is used to evaluate the effectiveness of new treatments by comparing the survival times of treatment and control groups. It's also used in engineering for reliability analysis of products, in finance for credit risk modeling, and in customer analytics for predicting churn rates. By accommodating censored data and providing tools for risk and time-to-event analysis, survival analysis offers a powerful framework for understanding and predicting scenarios where time plays a crucial role.

In medical research, survival analysis is extensively used to study the time until the occurrence of health-related events, offering valuable insights into patient outcomes and treatment efficacy \citep{ramanan2019comparison, kim2019effect, chen2020differential, jeon2021developing, ramanan2021impact, kim2021effect}. One common application is in clinical trials, where survival analysis is employed to compare the lifespans of patients receiving different treatments, helping to evaluate the effectiveness of new drugs or therapeutic approaches. It's instrumental in oncology research for analyzing time-to-remission or progression of disease, providing crucial information for treatment planning and patient counseling. Survival analysis is also pivotal in epidemiology for studying the factors affecting disease incidence and progression, which can include genetic, environmental, and lifestyle factors. Additionally, it's used in public health for analyzing patterns and determinants of mortality rates, thereby informing healthcare policy and preventive strategies \citep{jeon2021assessing}. The ability of survival analysis to handle censored data, where patients might be lost to follow-up or the study ends before all events have occurred, makes it particularly suited for longitudinal medical studies where follow-up duration varies among patients. This methodology helps in deriving more accurate and meaningful conclusions about patient survival and the factors influencing it, guiding clinical decision-making and healthcare interventions.

\subsection{Causal Inference}
Causal inference is a fundamental aspect of statistical analysis and research, focusing on determining the cause-and-effect relationships between variables \citep{rubin2005causal, pearl2009causal, pearl2010foundations, rosenbaum2017observation}. It goes beyond mere association or correlation, aiming to understand how a change in one variable directly affects another. 

Researchers often rely on observational data. Here, regression analysis, particularly multiple regression, is used to control for confounding variables. However, it's crucial to understand that correlation does not imply causation, and observational studies have limitations in establishing causality. In observational studies, propensity score is a technique used to reduce selection bias by matching treated and control subjects with similar characteristics \citep{rosenbaum1983central, imai2004causal, chen2021estimating}.

Causal inference is crucial across many disciplines, including epidemiology, economics, social sciences, and policy analysis. It helps in understanding the impact of interventions, policies, or treatments, and is key to making informed decisions based on empirical evidence. However, establishing causality is complex and requires careful consideration of the data, methodology, and underlying assumptions. Some techniques, like multiple imputation \citep{westreich2015imputation, chen2021estimating2}, instrumental variables \citep{baiocchi2014instrumental}, are often used in the causal inference. 

\subsection{High-Dimensional Data Analysis}
Dimensionality reduction techniques, such as Principal Component Analysis (PCA) \citep{ivanovic2015modern} and t-Distributed Stochastic Neighbor Embedding (t-SNE) \citep{cieslak2020t} are essential for reducing the complexity of genomic data, making it more manageable for analysis. Comparing with Network analysis, it helps in understanding the interrelationships between different genes or proteins and their role in diseases \citep{mcgillivray2018network,su2020network}.

High-dimensional data analysis is a critical area in statistics and machine learning, dealing with datasets that have a large number of variables (dimensions). This high dimensionality poses unique challenges, including increased computational complexity and the risk of overfitting models to the data. To address these issues, dimensionality reduction techniques are employed.

\subsection{Bayesian Methods and Machine Learning}
Bayesian analysis is a statistical method that applies the principles of probability to infer uncertainty in a wide range of applications \citep{abrams1994simple, verma2019application}. It's named after Thomas Bayes, an 18th-century mathematician, and it differs fundamentally from the classical frequentist approach to statistics.

Bayesian inference involves updating the probability of a hypothesis as more evidence or information becomes available. It combines prior knowledge (prior probabilities) with new data (likelihood) to form an updated probability (posterior probability).

Bayesian methods are particularly useful in personalized medicine for calculating individualized disease risks based on genetic information \citep{moon2007ensemble,voelzke2013personalized}. It has gained popularity with the advent of more powerful computers and sophisticated algorithms, allowing for its application in complex models that were previously computationally infeasible. Its ability to integrate prior knowledge and its flexibility in updating to reflect new information make it a powerful tool in statistical inference and decision-making.

Machine learning is a field of artificial intelligence that focuses on building systems capable of learning from and making decisions based on data \citep{jordan2015machine}. It involves algorithms and statistical models that enable computers to perform tasks without explicit instructions, relying instead on patterns and inference. This capability is achieved through a variety of methods, including supervised learning (where the system learns from labeled training data to make predictions or decisions), unsupervised learning (where it identifies patterns in unlabeled data), and reinforcement learning (where it learns to make decisions by receiving rewards for actions). Machine learning applications are incredibly diverse, especially in medical research \citep{zhou2022elucidation, zhou2022development, zhou2022f60p}. The application of machine learning in medical research represents a significant leap forward in diagnosing, treating, and understanding various health conditions. In this field, machine learning algorithms analyze complex medical data, which can include patient records, imaging data, genetic information, and even notes from healthcare providers, to uncover insights that might be too subtle or complex for traditional analysis.

\subsection{Clinical Trial Data Analysis}
The analysis of clinical trials is a critical aspect of medical research, providing the evidence base for evaluating the safety and efficacy of new treatments, drugs, and medical interventions. Clinical trials are systematic investigations in human subjects, designed to discover or verify the effects of one or more health-related interventions \citep{friedman2015fundamentals}. Trials are designed around specific endpoints, which could be primary (main outcomes of interest) or secondary (additional outcomes). These could include clinical endpoints like mortality or biomarkers.

Data from clinical trials are analyzed to assess the effectiveness and safety of the intervention.
Common statistical methods include comparison of means (using t-tests, ANOVA), survival analysis, regression models and historical data borrowing \citep{quan2022assessments}.
Intention-to-treat (ITT) analysis includes all randomized patients in the groups to which they were randomly assigned, regardless of whether they completed the treatment, ensuring unbiased comparisons.

The analysis of clinical trials is a sophisticated and highly regulated process \citep{levine1988ethics, califf2015exploring, abou2016deciphering, quan2022generalized}. It is essential for advancing medical science, informing clinical practice, and ensuring that new treatments are both effective and safe.

\section{Ethical and Privacy Considerations}
The utilization of large-scale patient data in precision medicine raises significant concerns regarding data security and patient privacy \citep{hulsen2019big}. With the increased collection of sensitive genetic, health, and lifestyle information, the risk of data breaches and unauthorized access becomes a paramount concern. The storage and transmission of vast amounts of personal health information necessitate robust cybersecurity measures. Data breaches can lead to sensitive information falling into the wrong hands, causing potential harm to individuals. The process of informed consent for patients whose data is being used in precision medicine research is complex. It must encompass the scope of the research, potential data sharing, and future uses of the data, ensuring that patients are fully aware of how their information will be utilized. To protect patient privacy, it's essential to de-identify personal health information. However, the challenge lies in effectively anonymizing data while retaining its utility for research purposes.

Predictive analytics in precision medicine, while beneficial, brings forth ethical challenges, particularly concerning the interpretation and application of these predictions \citep{ginsburg2018precision}. There is a potential risk that genetic information could be used to discriminate against individuals in areas like employment or insurance. The knowledge of one's genetic predisposition to certain diseases can have significant psychological impacts. Ethical considerations must include the management of this information and the provision of appropriate counseling. The use of predictive analytics must respect patient autonomy. Patients should be empowered to make informed decisions about their treatment plans based on predictive data, without coercion or undue influence \citep{molina2020impact}.

The development of comprehensive regulatory frameworks and guidelines is crucial in addressing these ethical and privacy concerns. National and international policies need to keep pace with technological advancements in precision medicine. These policies should govern data collection, storage, usage, and sharing. The adoption of standardized protocols for data handling and patient consent across research institutions and healthcare providers is necessary to ensure uniformity in ethical practices. As precision medicine evolves, so should the ethical frameworks and guidelines governing it. This requires continuous evaluation and adaptation of policies to address new ethical challenges as they arise.

\section{Future Directions and Emerging Trends}

The integration of artificial intelligence (AI) and machine learning (ML) in precision medicine represents a significant trend, with profound implications for future healthcare \citep{ahmed2020artificial}. AI and ML are at the forefront of developing sophisticated predictive models. These models can analyze vast and complex datasets, such as whole-genome sequences, to predict disease risks, patient responses to various treatments, and potential side effects with unprecedented accuracy \citep{delanerolle2021artificial}. AI algorithms are increasingly being tailored for personalized treatment. By analyzing patient data in real-time, these algorithms can suggest modifications in treatment plans, catering to the dynamic nature of individual patient responses. Cross-disciplinary AI Applications: AI's application is expanding beyond traditional boundaries, integrating insights from genomics, pharmacology, and patient-reported outcomes to devise comprehensive treatment strategies.

Precision medicine's future is also closely tied to advancements in genomic and molecular analysis. The integration of detailed genomic data with traditional clinical markers is paving the way for more nuanced and effective treatment strategies \citep{van2013clinical}. This synergy allows for a deeper understanding of the interplay between genetics, environment, and disease. Techniques like CRISPR-Cas9 \citep{jiang2017crispr} are revolutionizing the possibilities in precision medicine, offering potential cures for previously untreatable genetic diseases. Ongoing research in gene editing promises to open new avenues in disease management and prevention.

Digital Health and the Internet of Things (IoT) are emerging as key players in the future of precision medicine. Wearable health devices and remote monitoring tools are transforming patient care by providing continuous health data. This constant stream of data offers valuable insights into patient health, enabling proactive and preventive healthcare strategies. The proliferation of digital health tools generates massive amounts of health data. Leveraging big data analytics to derive actionable insights from this data is a growing trend, with significant potential to enhance patient care and research \citep{johnson2021precision}. The rise of telemedicine and virtual care platforms is another trend reshaping healthcare. These platforms extend the reach of precision medicine by providing access to specialized care and personalized treatment across geographical barriers.

As precision medicine advances, it brings complex ethical, legal, and social questions to the forefront. The ethical challenges surrounding the use of genomic data, including consent, privacy, and data ownership, are increasingly pertinent. Addressing these issues is crucial for the ethical advancement of precision medicine.
There is a growing focus on ensuring that the benefits of precision medicine are equitably distributed. Addressing disparities in access to precision medicine services is a critical area of focus to avoid exacerbating existing healthcare inequalities. The rapid pace of technological advancements in precision medicine presents regulatory challenges. Developing flexible yet robust regulatory frameworks that can adapt to the evolving landscape of precision medicine is vital for its sustainable and ethical growth.

\section{Case Studies in Precision Medicine}
Several case studies highlight the successful application of data integration techniques in precision medicine. In cancer research, the integration of genomic data with clinical outcomes has led to the identification of biomarkers for cancer prognosis and the development of targeted therapies.
In neurodegenerative diseases, studies combining genetic, imaging, and cognitive data have provided insights into the pathophysiology of diseases like Alzheimer’s, aiding in the development of diagnostic tools and therapeutic strategies. Integration of EHRs with genomic data has enabled the creation of personalized treatment plans in various medical fields, improving patient outcomes and treatment efficiency.
\subsection{Oncology}
The field of oncology has been one of the primary beneficiaries of precision medicine, with significant advancements in the diagnosis and treatment of various cancers.
The use of genomic profiling has revolutionized cancer treatment. By analyzing the genetic makeup of tumors, clinicians can identify specific mutations and select targeted therapies that are more effective and have fewer side effects compared to traditional chemotherapy.

The identification of predictive biomarkers through statistical analysis has led to the development of personalized therapies. For instance, the use of HER2-targeted therapies in breast cancer patients with HER2-positive tumors has significantly improved outcomes \citep{singh2014her2,oh2020her2}.

Statistical models have also been instrumental in predicting patient responses to immunotherapies, a breakthrough treatment modality in oncology. Machine learning algorithms analyze patient and tumor characteristics to identify those who are most likely to benefit from immunotherapy.

\subsection{Cardiovascular Diseases}
Precision medicine is making strides in the field of cardiovascular diseases, offering new insights into risk assessment and management \citep{currie2018precision}.

The development of genetic risk scores for cardiovascular diseases is a prime example of precision medicine. These scores, derived from statistical analysis of genetic data, help in assessing an individual's risk of developing heart disease. Statistically-driven analyses have led to the identification of genetic variants that affect individual responses to cardiovascular drugs, such as statins \citep{franks2023precision}. This has enabled the personalization of drug therapies, enhancing their efficacy and safety.

\subsection{Neurological Disorders}
Precision medicine is also impacting the field of neurology, particularly in the management and understanding of neurological disorders \citep{forloni2020alzheimer}.

In Alzheimer's disease, precision medicine approaches are being used to identify early biomarkers of the disease through statistical analysis of genetic \citep{hampel2020path}, imaging, and cognitive data. This early detection is crucial for timely intervention \citep{berkowitz2020precision}.

The use of precision medicine in epilepsy has led to personalized treatment approaches. Statistical analysis of patient data helps in predicting which patients are likely to respond to specific antiepileptic drugs, thereby optimizing treatment.

\section{Discussion}
This review has explored the multifaceted role of statistical analysis in the evolving field of precision medicine. We have seen how advanced statistical techniques are essential for interpreting complex, multidimensional datasets, enabling the transition from traditional healthcare approaches to personalized medical treatments. Key areas, such as predictive modeling, high-dimensional data analysis, and Bayesian methods, have been pivotal in this transition.

In oncology, cardiovascular diseases, and neurological disorders, among others, precision medicine has demonstrated remarkable potential in improving patient outcomes through targeted therapies and personalized treatment plans. The integration of various data types, including genomic, clinical, imaging, and lifestyle information, has been a crucial aspect of this evolution, despite the challenges in data heterogeneity and integration.

The vast and intricate nature of datasets in precision medicine poses significant challenges for data analysis and interpretation. One of the main challenges is dealing with high-dimensional data, such as genomic or proteomic data, which often contains more variables than observations. This dimensionality poses problems for traditional statistical methods and increases the risk of overfitting and spurious correlations. Integrating diverse types of data, such as clinical records, imaging, and molecular data, is another major challenge. The heterogeneity in formats, scales, and sources of these data requires sophisticated integration techniques to ensure accurate and meaningful analysis. Precision medicine datasets often contain missing or incomplete data, which can significantly bias the results of the analysis. Developing robust methods to handle missing data is crucial for the reliability of statistical models in precision medicine.

The balance between overfitting and underfitting is a critical issue in the development of predictive models in precision medicine. Overfitting occurs when models are too complex and capture noise instead of underlying patterns, leading to poor performance on new data. On the other hand, underfitting happens when models are too simple to capture the complexity in the data, resulting in inaccurate predictions. Techniques such as cross-validation and regularization are employed to prevent overfitting and underfitting \citep{ghojogh2019theory}. Cross-validation involves dividing the data into subsets to validate the model's performance, while regularization methods add a penalty to the model complexity.

The trade-off between model complexity and interpretability is a central concern in precision medicine. While complex models, like deep learning algorithms, can provide high accuracy, their "black-box" nature makes it difficult to understand how they make predictions. This lack of interpretability can be a significant barrier in clinical settings where understanding the rationale behind a decision is crucial. There is a growing emphasis on developing explainable AI (XAI) \citep{dwivedi2023explainable} approaches that offer transparency in how models arrive at their predictions. XAI is essential for gaining the trust of clinicians and patients and for the ethical application of AI in healthcare.

The impact of statistical analysis on the future of precision medicine is profound and far-reaching. As we continue to amass larger and more diverse datasets, the need for sophisticated statistical methodologies becomes increasingly critical. These methodologies not only enhance our understanding of complex diseases but also guide the development of more effective and personalized treatments.

The integration of AI and machine learning in precision medicine offers exciting possibilities for future healthcare. These technologies promise to further refine predictive models, making them more accurate and efficient in identifying suitable treatments for individual patients.

However, as we advance, we must also be mindful of the challenges and limitations. The balance between model complexity and interpretability, ethical considerations, and data privacy concerns are crucial factors that need continuous attention. Additionally, the field must address the issues of data quality and standardization to fully realize the potential of precision medicine.

Ethical and regulatory considerations are integral to the advancement of precision medicine. Ensuring patient consent and maintaining data privacy are paramount. As the use of patient data becomes more complex, so does the responsibility to protect it and use it ethically. Navigating the regulatory landscape, which is often lagging behind technological advancements, poses a challenge. Ensuring that new methodologies and technologies comply with existing regulations while advocating for updated policies is crucial.

Looking forward, the field of precision medicine is set to revolutionize healthcare, with statistical analysis playing a key role in this transformation. By continuing to develop and refine statistical methodologies, and by addressing the associated ethical and practical challenges, precision medicine can provide more accurate, effective, and personalized healthcare solutions for patients worldwide.

\bibliography{mybib}

\end{document}